\documentclass[conference]{IEEEtran}
\IEEEoverridecommandlockouts
\usepackage{cite}
\usepackage{amsmath,amssymb,amsfonts}
\usepackage{algorithmic}
\usepackage{graphicx}
\usepackage{textcomp}
\usepackage{multirow}
\def\BibTeX{{\rm B\kern-.05em{\sc i\kern-.025em b}\kern-.08em
    T\kern-.1667em\lower.7ex\hbox{E}\kern-.125emX}}
\begin{document}

\title{DeepUNet: A Deep Fully Convolutional Network for Pixel-level Sea-Land Segmentation\\
\thanks{This work was supported by the National Natural Science Foundation of China under Grant No.~$91638201$, Grant No.~$61501018$, Grant No.~$61571033$, and the Higher Education and High-Quality and World-Class Universities (PY201619).}
}

\author{
\IEEEauthorblockN{Ruirui Li, Wenjie Liu, Lei Yang, Shihao Sun,\\ Wei Hu*\thanks{* The corresponding author.}, Fan Zhang, \textit{Senior Member, IEEE}, Wei Li, \textit{Senior Member, IEEE}}
\IEEEauthorblockA{\textit{Beijing University of Chemical Technology}\\
Beijing, China \\
ilydouble@gmail.com, 1186894139@qq.com, ylxx@live.com, 472527311@qq.com,\\ huwei@mail.buct.edu.cn, zhangf@mail.buct.edu.cn, liw@mail.buct.edu.cn}
}

\maketitle

\begin{abstract}

Semantic segmentation is a fundamental research in remote sensing image processing. Because of the complex maritime environment, the sea-land segmentation is a challenging task.  Although the neural network has achieved excellent performance in semantic segmentation in the last years, there are a few of works using CNN for sea-land segmentation and the results could be further improved. This paper proposes a novel deep convolution neural network named DeepUNet. Like the U-Net, its structure has a contracting path and an expansive path to get high resolution output. But differently, the DeepUNet uses DownBlocks instead of convolution layers in the contracting path and uses UpBlock in the expansive path. The two novel blocks bring two new connections that are U-connection and Plus connection. They are promoted to get more precise segmentation results. To verify our network architecture, we made a new challenging sea-land dataset and compare the DeepUNet on it with the SegNet and the U-Net. Experimental results show that DeepUNet achieved good performance compared with other architectures, especially in high-resolution remote sensing imagery.

\end{abstract}


\section{Introduction}

Optical remote sensing images play an important role in maritime safety, maritime management, and illegal smuggling as they can provide more detailed information compared with SAR images. For remote sensing imagery, sea-land segmentation is aimed to separate coastline or wharf images into ocean region and land region, which is of great significance to ship detection and classification, since a clear coastline can reduce the number of ships in the wrong mark.\\

The sea-land segmentation task is very challenging. First of all, the interference of the atmospheric factors could not be neglected. These factors include clouds, shadows, waves caused by the wind, and etc. Traditional thresholding based methods such as OTSU\cite{b1}, LATM\cite{b2} often fails due to the complicated distribution of intensity and texture. Secondly, images containing the marine and terrestrial environment are of complex semantic contents. There probably are ships, inland waters, islands, and forests that confusing the algorithms. As a result,  early learning based methods cannot solve the problems of misclassification.  In the last several years, convolutional neural networks (CNNs) have been widely developed in computer vision and semantic segmentation. For sea-land segmentation task, the SeNet\cite{b3} has been proposed, which combines the segmentation task and edge detection task into an end-to-end deconvNet\cite{b4} in a multi-task way. The SeNet also promoted the Local Regularized loss to decrease the misclassification. In their experiments, the SeNet achieved better results than original deconvNet. In fact, remote sensing images are usually of high resolution, for example, 1024$\times$1024 or 2048$\times$2048. They contain both large areas and small targets in one image, which require deeper convolutional network to take both the high-level global features and the low-level local features into considerations.\\

To solve this problem, in this paper we explored a novel network structure named DeepUNet for pixel-level sea-land segmentation. DeepUNet is an end-to-end fully convolutional network with two other kinds of short connections. We call them U connections and Plus connections.  The main idea of the DeepUNet is to concatenate the layers in the contracting path to that of  expansive path. High-resolution features from the contracting path are combined with the upsampled output. Hence a successive convolution layer can then learn to assemble a more precise output based on this information.  Furthermore, to better extract high-level semantic information with less loss error, the proposed DeepUNet optimize the contracting path as well as the expansive path by introducing the DownBlock, the UpBlock and the Plus connections. In the DownBlock and the UpBlock, features before and after convolution layers are added together. This structure skips the invalid convolution operation and supplies a deeper and efficient convolution neural architecture.\\

To prove the DeepUNet in sea-land segmentation, we collected images from different places and of various illuminated condition from the google earth (GE). With the handicraft labeled ground truth images (GT), we compare the U-Net, SegNet and DeepUNet on the collected dataset. Experiments demonstrated that the DeepUNet achieve high precision-recall and F1-measure for both sea and land regions.\\

In summary, this paper makes the following contributions to the community:
\begin{itemize}
\item A new dataset for sea-land segmentation is provided. It contains 207 handicraft labeled images in which 122 for training, and 85 images for validating.
\item A novel convolutional network structure is introduced for remote-sensing image segmentation, named DeepUNet.  It is concise and efficient. Compared with other architecture, it gets better sea-land segmentation results.
\item We perform a complete comparison among SegNet, U-Net, and DeepUNet on the provided dataset.\\
\end{itemize}

The remainder of the paper is organized as follows. The section 2 reviews related works and differentiates our method from such works. The Section 3 and Section 4 introduce our proposed method as well as detailed implementations.  Extensive experimental results and comparisons are presented in section 5. And section 6 concludes this paper.

\section{Related Works}

\subsection{Sea-land segmentation}

Sea-land segmentation has been a hot area for remote sensing image processing. For multispectral imagery, the map of normalized difference water index (NDWI) is often calculated in the near-infrared (NIR) band to enhance the water areas while suppressing the green land and soil areas. These works include Kuleli et al.\cite{b5}, Di et al.\cite{b6}, Zhang et al.\cite{b7}, Aktas et al.\cite{b8}, Aedla et al.\cite{b9}, and etc.\\

However, for natural-colored imagery, there is limited literature for sea-land segmentation and coastline extraction. Most of the existing works are based on thresholding algorithms, accompanied with morphological operations to eliminate errors in the results. For example, Liu\cite{b2} proposed an automatic threshold determination algorithm for the local region. You and Li\cite{b10} built a Gaussian statistical sea model based on the OTSU (Otsu 1979)\cite{b1}.  Zhu et al. (Zhu et al. 2010)\cite{b11} enhanced the images first and segmented the enhanced images using the OTSU as well. The OTSU algorithm is considered to make the optimal threshold selection in image segmentation, which is not affected by image brightness and contrast, so it has good performance in simple sea-land segmentation tasks. The thresholding based methods only employ the spectral information of individual pixel and ignore the local relationship of neighboring pixels. The results of them often contain misclassiﬁcation, especially in the land area. To solve the problem, the learning based methods are proposed. They try to extract the features of local small areas in remote sensing imagery and train the weight of these features to classify the sea and land. Xia\cite{b12} integrates LBP feature to sea-land segmentation. Cheng\cite{b13}  clustered the pixels into super-pixels and promoted a super-pixel based seeds learning for sea-land segmentation. These learning based methods rely on the manually selected features in a large degree. As a result, for remote sensing imagery with complex semantic information, these methods also have plenty of misclassified pixels.   For instance, the shadow and green colored regions in land areas may be classiﬁed as water, while waves and noises in water areas may be considered as land.\\

Recent advancement in deep learning motivates researchers to address these problems with deep neural networks.  Two states of the art works have been found. The last sea-land segmentation algorithm is SeNet\cite{b3} which is based on DeconvNet framework. The SeNet designed a multi-task way, thus it can do sea-land segmentation and edge detection at the same time.  Lin et al.\cite{b14} proposed a multi-scale fully convolutional network for maritime semantic labeling. They divide the pixels of maritime imagery into three classes that are sea, land, and ships. Because of the pooling steps of the FCN, the output of the images cannot provide high-resolution segmental results.  Despite lots of efforts they did, challenges on remote sensing image segmentation are far from resolved.  Currently, neither the SeNet nor the multi-scale structure network is intelligent enough to segment well, especially in the case of high-resolution remote sensing imagery with plenty of detailed contents.\\

\subsection{Deep learning for semantic segmentation}

Semantic segmentation is aimed to understand an image in pixel level. Its main task is to label each pixel into a certain class.  Deep learning based object detection and semantic segmentation in computer vision have made a big advancement. The Fully Convolutional Networks (FCNs)\cite{b15}, proposed by Long et al. from Berkeley, is a landmark in image segmentation. It first allowed pixel-level segmentation by replacing fully connected neural layers with convolutional neural layers.However, the FCNs produce coarse segmentation maps because of the loss of information during pooling operations. Thus lots of research focuses on how to provide pixel-level high-resolution segmentation results. There are two kinds of works addressing this problem.\\

The first kinds are based on dilated convolution\cite{b16} (also called as atrous convolution). Lots of works are proposed to improve the dilated convolution including atrous spatial pyramid pooling\cite{b17}, fully connected CRF\cite{b18} and etc..\\

Other efforts are made to build connections between the pooling layers and the unpooling layers. In the convolutional network, the max pooling operation is non-invertible; however, we can obtain an approximate inverse by recording the locations of the maxima within each pooling region in a set of switch variables. For example, in the DeconvNet, the unpooling operation uses the switches to place the reconstructions from the layer above into appropriate locations, preserving the structure of the stimulus. The SegNet is very like the DeconvNet in structure but is different in the unpooling strategy. The SegNet\cite{b19} proposed an encoder-decoder convolutional network which  transfers encoded maxpooling indices to decoder to improve the segmentation resolution. Another impressive CNN structure is U-Net\cite{b20} which is proposed for biomedical image segmentation.  Its architecture consists of a contracting path and an expansive path and its feature maps from the contracting path are cropped and copied for the correspondingly upsamplings in the expansive path. Inspired by the U-Net architecture, our work supplies two connections that are u-connection and plus connection.  It replaces the contracting path and the expansive path with successive DownBlock and UpBlock which are described in detail in Section 3.

\section{Proposed method}

With the improvement of the spatial resolution of satellite and aircraft sensors, more details of the intensity and texture are presented in remote sensing images, which makes the segmentation problem more challenging.  On the other hand, for image classification in computer vision, deeper networks are proved to be able to get better accuracy and thus become popular.  Currently, both the last two CNN-based methods for sea-land segmentation are based on VGG16 structure. Through multi-task techniques and multi-scale techniques, they alleviate the problem of misclassification. But they probably fail when facing images with rich semantic information. Here, we proposed the DeepUNet which is specifically designed for high resolution images with detailed contents and objects. This network has the reception field covering the whole image while has the ability to distinct the small area in the images as Fig.1 illustrated. Our network structure does not conflict with the last two CNN-based works and can be further improved combined with multi-task techniques.

\begin{figure}[htbp]
\setlength{\abovecaptionskip}{0pt}
\setlength{\belowcaptionskip}{-10pt}
\centering
\includegraphics[width=.5\textwidth]{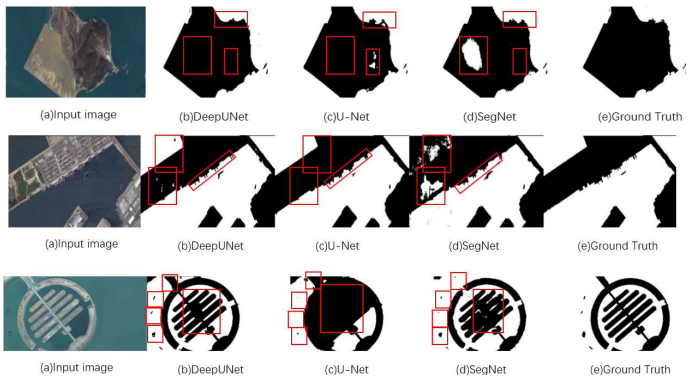}
\caption{the comparison of the DeepUNet,U-Net,SegNet and the Ground truth}
\label{fig1}
\end{figure}

This section begins with the main framework of the DeepUNet, which introduces the basic idea and the architecture first. Then we describe the DownBlock and the UpBlock in detail with which together greatly enhance the performance of the network when dealing complex segmentation tasks.

\subsection{Network Structure}
The process of the DeepUNet is simply illustrated in Fig.2. It provides an end-to-end network. The input images are three channels’ RGB remote sensing images and the output images are binary segmentation maps in which the white pixels symbolize the sea and the land vice versa. The network does not have any fully connected layers and only uses 1x1 convolution layer for dimension reduction. At the end, we use a Softmax layer to transform the results of the neural network into a two-class problem. This strategy allows the seamless segmentation of arbitrarily large images by an overlap-tile strategy.\\

The structure defines two kinds of blocks. In Fig.2, the blue bold bar is named DownBlock, and the green bold bar is name UpBlock. Like the U-Net, our structure is symmetrical. The left side path consists of repeated DownBlocks which are connected to the corresponding UpBlocks. This connection is showed with the yellow lines in the Fig.2. We called them u-connections since they concatenate the feature maps of the DownBlock to that of the corresponding UpBlock. Besides the u-connection , there is another kind of short connections between the successive DownBlocks or UpBlocks. They are showed with purple lines in the Fig.2 and called the Plus connection or the Plus layer. The Plus layer is an optimized structure. It can solve the problem that the loss error increases when the network goes deep. The Plus layer avoids the training step converge on the local optimal solution and thus guarantees the very deep networks achieve good performance in complex image segmentation task.

\begin{figure*}[htbp]
\setlength{\abovecaptionskip}{0pt}
\setlength{\belowcaptionskip}{-10pt}
\centering
\includegraphics[width=.95\textwidth]{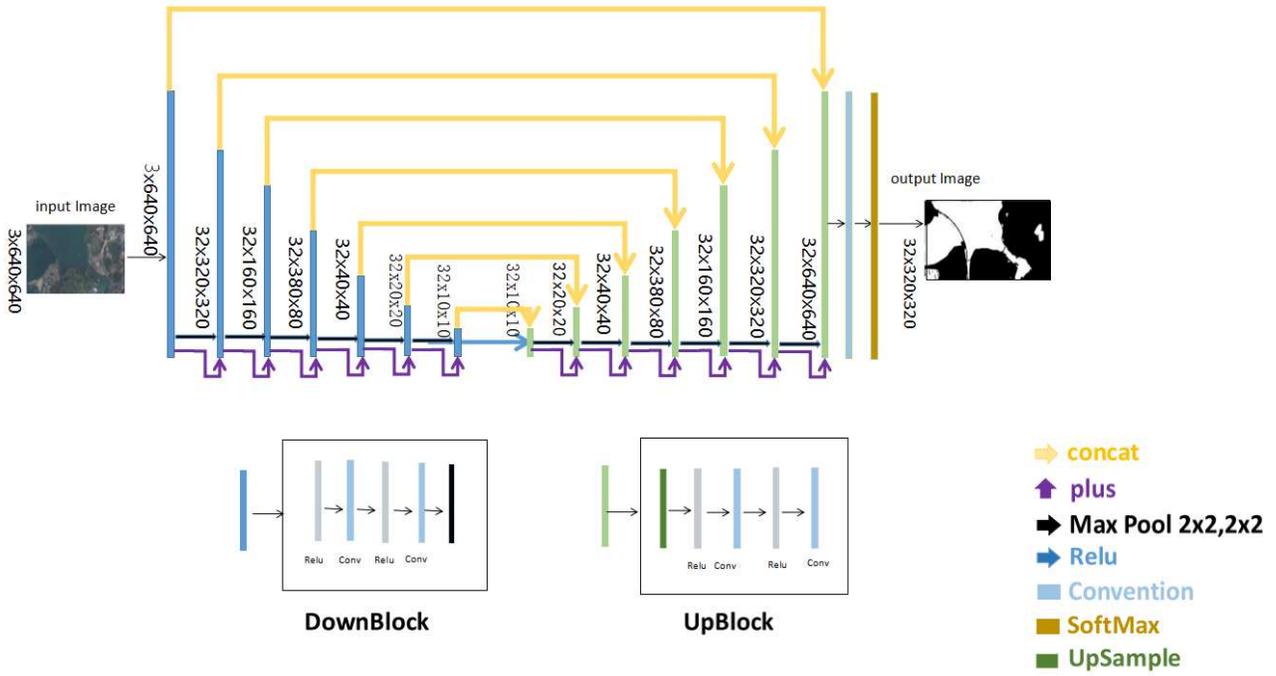}
\caption{DeepUNet detailed structure and annotations}
\label{fig2}
\end{figure*}

\subsection{Down-sampling block}

The DownBlock has two convolutional layers that are concatenated through the ReLU layer. The first convolutional layer uses a 3$\times$3 convolution kernel, a 1$\times$1 step size, and a total of 64 convolution cores. The second layer uses a 3$\times$3 convolution kernel, a 1$\times$1 step size, and a total of 32 convolution cores. The DeepUNet chooses two convolutions of small kernel size instead of a larger single convolution kernel. The reception field of two successive 3$\times$3 convolutional layers is the same with that of a 5$\times$5 convolutional layer; but in the former choice, the parameters that have to be computed are much less.\\

\begin{figure}[htbp]
\setlength{\abovecaptionskip}{0pt}
\setlength{\belowcaptionskip}{-10pt}
\centering
\includegraphics[width=.45\textwidth]{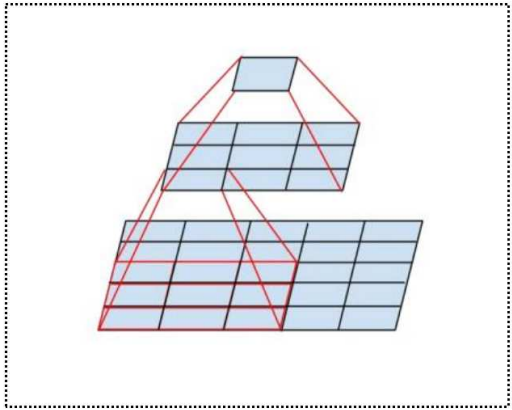}
\caption{reception field of two successive 3$\times$3 convolutional layers}
\label{fig3}
\end{figure}

The input of the block is 32-dimension features. It is of the same feature size with that of the second convolution layer’s output. A Plus layer is added after the convolution operation. 

\begin{figure}[htbp]
\setlength{\abovecaptionskip}{0pt}
\setlength{\belowcaptionskip}{-10pt}
\centering
\includegraphics[width=.45\textwidth]{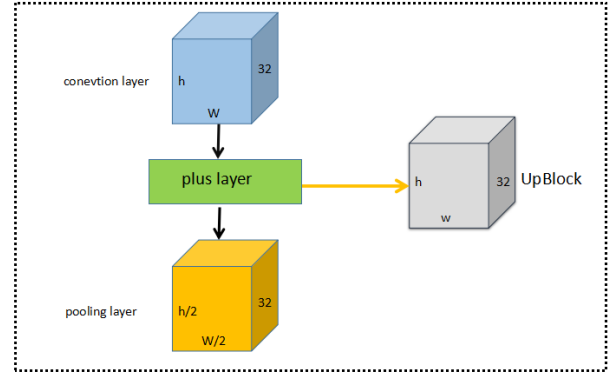}
\caption{DownBlock results send to UpBlock and next DownBlock}
\label{fig4}
\end{figure}

Assuming $y$ is the output of the Plus layer, $x$ is the input of the DownBlock, the Plus layer passes $x$ and the result of the second convolution layer through the (1), and leaves optimal results $y$ into max pooling layer. \\

\begin{figure}[htbp]
\setlength{\abovecaptionskip}{0pt}
\setlength{\belowcaptionskip}{-10pt}
\centering
\includegraphics[width=.45\textwidth]{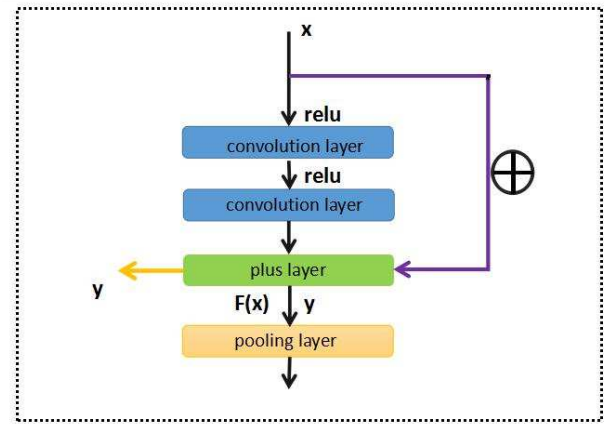}
\caption{Detail of DownBlock}
\label{fig5}
\end{figure}

In the (1), $W_{1}$ symbolizes the first convolution operation, $W_{2}$ symbolizes the second convolution operation, and $\sigma$ illustrates the ReLU function.

\begin{equation}
y=W_{2}\sigma(W_{1}x)+x
\end{equation}

The max pooling layer in the DownBlock has a kernel size of 2$\times$2 and a step size of 2$\times$2. Here, we not only pass $y$ to the max pooling layer; but also concatenate the feature maps to the corresponding UpBlock of the same level.

\subsection{Up-sampling Block}
The DeepUNet adopts an elegant architecture that is symmetric. The UpBlock is promoted to assemble a more precise output. The structure of the module is basically the same as that of the DownBlock module (Fig.5).

\begin{figure}[htbp]
\setlength{\abovecaptionskip}{0pt}
\setlength{\belowcaptionskip}{-10pt}
\centering
\includegraphics[width=.45\textwidth]{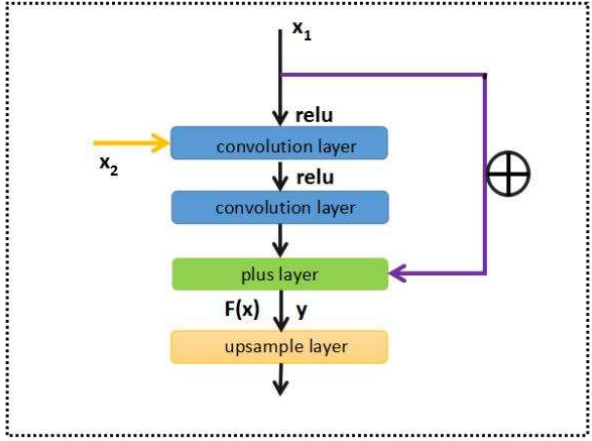}
\caption{Detail of UpBlock}
\label{fig6}
\end{figure}

It also contains two convolutional layers and a Plus layer.  But differently, there is an upsampling layer instead of the max pooling layer. The input of the convolution layer is a concatenated feature map named $x$  

\begin{equation}
x=[\delta,x_{1},x_{2}]
\end{equation}

In the (2), $x_{1}$ is the feature map from the previous UpBlock and $x_{2}$ is that from the DownBlock through u-connection and $\delta$ is the upsampling operator. On the basis of the structure, the DeepUNet passes the information before max pooling to the same level of the UpBlock through the connected channel. The information is processed by the convolutional layers and is helped to get more precise results during the upsampling. According to the architecture, we have to keep the resolution of DownBlock’s output the same with that of UpBlock’s input. Thus we add the upsampling layer in the beginning of the block.\\

In summary, the detailed parameters of DeepUNet layers are listed in Table \uppercase\expandafter{\romannumeral1}

\begin{table}[h]
\caption{the detailed parameters of DeepUNet layers}
\newcommand{\tabincell}[2]{\begin{tabular}{@{}#1@{}}#2\end{tabular}}
\centering
	\begin{tabular}{|c|c|c|c|}
	\hline
	Layer name & Kernel size & Kernel Number & Remark \\
	\hline
	\tabincell{c}{conv0\_0\\conv0\_1\\conv0\_2\\Pooling0} & \tabincell{c}{3$\times$3\\3$\times$3\\2$\times$2\\2$\times$2} & \tabincell{c}{64\\64\\32\\-} & \multirow{21}*{Down pooling  stage} \\
	\cline{1-3}
	\tabincell{c}{conv1\_1\\conv1\_2\\Pooling21} & \tabincell{c}{3$\times$3\\2$\times$2\\2$\times$2} & \tabincell{c}{64\\32\\-} & \\
	\cline{1-3}
	\tabincell{c}{conv3\_1\\conv3\_2\\Pooling31} & \tabincell{c}{3$\times$3\\2$\times$2\\2$\times$2} & \tabincell{c}{64\\32\\-} & \\
	\cline{1-3}
	\tabincell{c}{conv4\_1\\conv4\_2\\Pooling41} & \tabincell{c}{3$\times$3\\2$\times$2\\2$\times$2} & \tabincell{c}{64\\32\\-} & \\
	\cline{1-3}
	\tabincell{c}{conv5\_1\\conv5\_2\\Pooling51} & \tabincell{c}{3$\times$3\\2$\times$2\\2$\times$2} & \tabincell{c}{64\\32\\-} & \\
	\cline{1-3}
	\tabincell{c}{conv6\_1\\conv6\_2\\Pooling61} & \tabincell{c}{3$\times$3\\2$\times$2\\2$\times$2} & \tabincell{c}{64\\32\\-} & \\
	\cline{1-3}
	\tabincell{c}{conv7\_1\\conv7\_2\\Pooling71} & \tabincell{c}{3$\times$3\\2$\times$2\\2$\times$2} & \tabincell{c}{64\\32\\-} & \\
	\hline
	\tabincell{c}{Upsample81\\conv8\_1\\conv8\_2} & \tabincell{c}{-\\3$\times$3\\3$\times$3} & \tabincell{c}{-\\64\\32} & \multirow{21}*{Up Sampling stage} \\
	\cline{1-3}
	\tabincell{c}{Upsample91\\conv9\_1\\conv9\_2} & \tabincell{c}{-\\3$\times$3\\3$\times$3} & \tabincell{c}{-\\64\\32} & \\
	\cline{1-3}
	\tabincell{c}{Upsample101\\conv10\_1\\conv10\_2} & \tabincell{c}{-\\3$\times$3\\3$\times$3} & \tabincell{c}{-\\64\\32} & \\
	\cline{1-3}
	\tabincell{c}{Upsample111\\conv11\_1\\conv11\_2} & \tabincell{c}{-\\3$\times$3\\3$\times$3} & \tabincell{c}{-\\64\\32} & \\
	\cline{1-3}
	\tabincell{c}{Upsample121\\conv12\_1\\conv12\_2} & \tabincell{c}{-\\3$\times$3\\3$\times$3} & \tabincell{c}{-\\64\\32} & \\
	\cline{1-3}
	\tabincell{c}{Upsample131\\conv13\_1\\conv13\_2} & \tabincell{c}{-\\3$\times$3\\3$\times$3} & \tabincell{c}{-\\64\\32} & \\
	\cline{1-3}
	\tabincell{c}{Upsample141\\conv14\_1\\conv14\_2} & \tabincell{c}{-\\3$\times$3\\3$\times$3} & \tabincell{c}{-\\64\\32} & \\
	\hline
	\end{tabular}
\label{table1}
\end{table}

\section{Implementation details}
\subsection{Data preprocessing}

Data augmentation is essential to teach the network the desired invariance and robustness properties, when only few training samples are available. In case of remote sensing images, we primarily need shift and rotation invariance as well as scale variations. The data for training are square images randomly cropped from the augmented data. To enhance the efficiency of the training, we only choose those cropped images containing both sea and land. In 122 high-resolution remote sensing images, we finally generate 24000 training samples.

\subsection{network definition}

We implement the DeepUNet on the mxNet\cite{b21}. The mxNet is an excellent deep learning framework that provides two ways to program: shallow embedded mode and deep embedded mode. We use the deep embedded mode to realize our idea.\\

Our network defines the convolutional layer, the ReLU layer, and the pooling layer through the sym model of the mxNet. The defined layers are then added to the UpBlock and DownBlock according to the network design.

\subsection{Training}

To minimize the overhead and make maximum use of the GPU memory, we favor large input tiles over a large batch size.  For Nvidia 1080Ti GPU, We choose 640$\times$640 square images and hence reduce the batch to 11 samples. The epoch that is number of learning steps is set to 10000. We use a high momentum (0.9) .The initial learning rate is 0.1, when the number of training steps reaches half of the total learning steps and the learning rate is adjusted to 0.01. When the number of training steps reaches 3/4 of the total learning step, set the learning rate as 0.001.\\

We set the Softmax function to sort out the results. The Softmax is the generalization of logistic function that converts all the results to probabilities between (0,1). In the task of sea-land segmentation, the DeepUNet divides the pixels into sea and land; thus the Softmax function $S_{i}$ is simplified by (3).

\begin{equation}
S_{i}=\frac{e^{v_{i}}}{\sum^k_{2} e^{v_{k}}}
\end{equation}

\subsection{Overlap tiles}

In the predicting step, we cropped the large image into  640$\times$640  tiles, and test the tiles one by one from bottom left to up right in a sliding window way.  The cropped step is not  necessary, but for high-resolution image we have to do it because of the limitation of GPU memory. We propose an overlap tiles strategy.  To predict the pixels in the border region of the image the missing context is extrapolated by mirroring the input image. For each tile, we compute the weight for overlap areas by the Gaussian function in which the distance is between current pixels and the center of the tile. Through weighted summary, we composite the overlap tiles and seamless  stitch the whole segmental image.

\section{Experiments and analysis}
\subsection{Experiments setup}

The experiments are carried out on a laboratory computer. Its configuration is shown in Table \uppercase\expandafter{\romannumeral2}. The operating system is installed of Ubuntu 16.04. The main required packages include python 2.7, CUDA8.0, cudnn7, tensorflow1.3.0, caffe, keras1.2.0, mxNet0.10.0 and etc.

\begin{table}[h]
\caption{experimental environments}
\centering
\begin{tabular}{|c|c|}
\hline
\textbf{CPU} & Intel (R) Core (TM) i7-4790K 4.00Hz \\
\hline
\textbf{GPU} & GeForce GTX1080 Ti \\
\hline
\textbf{RAM} & 20GB \\
\hline
\textbf{Hard disk} & Toshiba SSD 512G \\
\hline
\textbf{System} & Ubuntu 16.04 \\
\hline
\end{tabular}
\label{table2}
\end{table}

To prove the efficiency of the DeepUNet, we compare it with the U-Net and the SegNet using the same dataset and on the same experimental environment. The source code of U-Net and SegNet are all downloaded from the Github web pages that their authors provided. We train each model for all networks on the 122 high resolution images without any pertained model and test the models on the left 85 images to prove their generalization.\\

The deepUNet is developed for more complex sea-land segmentation as it can provide deeper convolutional structure with low loss error. To verify this point, in the experiments, we increase the resolution and complexity of the remote sensing images and analysis the results.

\subsubsection{Datasets preparation}

The dataset contains 207 remote sensing images which are collected from the Google Earth. Since we focus on sea-land segmentation, the images we selected are all from coastline and wharfs. We captured images by the software Google Earth provided and we chose viewpoints in space resolutions ranging from 3m to 50m. The satellites images we obtained are unlabeled, so we used the Photoshop to manually label the ground truth for all the images. Among them, 122 images were randomly selected as training sets, and the left 85 images were used for verification and testing. Our dataset has multi-scale images. Fig.7 shows some images collected from different heights but in the same location.

\begin{figure}[htbp]
\setlength{\abovecaptionskip}{0pt}
\setlength{\belowcaptionskip}{-10pt}
\centering
\includegraphics[width=.45\textwidth]{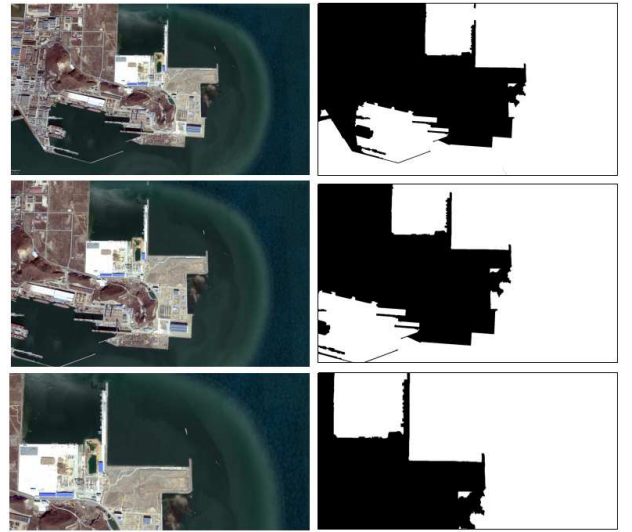}
\caption{images collected from different heights but in the same location}
\label{fig7}
\end{figure}

\subsubsection{Evaluation Metrics}
In the comparison experiments, we use Precision, Recall, F1metric to evaluate the proposed method. The sea-land segmentation task concerns not only the sea region but also the land region. In this paper, we calculated land precision (LP), land recall (LR), overall precision (OP), and overall recall (OR) which are defined as follows:

\begin{align}
& LP=\frac{TP_{L}}{TP_{L}+FP_{L}},LR=\frac{TP_{L}}{TP_{L}+FN_{L}} \\
& OP=\frac{TP_{L}+TP_{S}}{TP_{L}+FP_{L}+TP_{S}+FP_{S}} \\
& OR=\frac{TP_{L}+TP_{S}}{TP_{L}+FN_{L}+TP_{S}+FN_{S}} 
\end{align}

where TP(land), FP(land), and FN(land) are true positive, false positive, and false negative of land. TP(sea), FP(sea),and FN(sea) are true positive, false positive, and false negative of sea. OP combines precision of land and sea. OR combines recall of land and sea. The F1-measure is defined as,

\begin{equation}
F1=\frac{2\cdot Precision\cdot Recall}{Precision+Recall}
\end{equation}

\subsection{Comparison and Analysis}
We compare the U-Net, SegNet and DeepUNet on the same experimental environment. Part of the obtained results are shown in Figure 8, 9, 10, 1. In these figures we can clearly figure out that the proposed method is significantly outperformed the other methods.

\begin{figure}[htbp]
\setlength{\abovecaptionskip}{0pt}
\setlength{\belowcaptionskip}{-10pt}
\centering
\includegraphics[width=.45\textwidth]{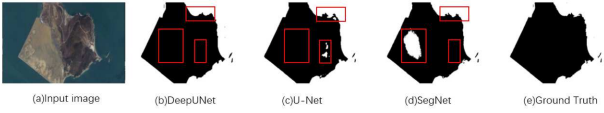}
\caption{The segmentation results of island predicted by DeepUNet,U-Net,SegNet}
\label{fig8}
\end{figure}

Fig.8(a) shows an optical image containing an island that almost covering the whole image. The island has uneven surface color because of the sunlight. Fig.8(b) shows the result of the DeepUNet. Compared with Fig.8(c) U-Net and Fig.8(d) SegNet, it completely segments the island without internal errors. The proposed network has a reception field of 4220$\times$4220 that is covering the image, and thus it takes the global features including the connectivity into considerations. For Fig.8 the evaluation table is listed in Table \uppercase\expandafter{\romannumeral3}. The indicators show that DeepUNet's OP is 3.65\% higher than U-Net and 3.26\% higher than SegNet. The F1-measure of DeepUNet is 4.8\% higher than U-Net and 1.92\% higher than SegNet.

\begin{table}[h]
\centering
\caption{the evaluation results of Fig.8}
\begin{tabular}{|c|c|c|c|c|c|}
\hline
Name & LP(\%) & LR(\%) & OP(\%) & OR(\%) & F1(\%) \\
\hline
DeepUNet & \textbf{98.90} & \textbf{99.76} & \textbf{99.41} & \textbf{99.41} & \textbf{99.32} \\
\hline
U-Net & 99.73 & 98.52 & 99.24 & 99.24 & 94.10 \\
\hline
SegNet & 99.69 & 91.72 & 96.25 & 96.25 & 95.53 \\
\hline
\end{tabular}
\label{table3}
\end{table}

For a harbor case, the results of different networks are shown in Fig.9. The test image not only contains small ships and shadows, but also contains grassland. These factors make the segmentation task difficult. From the Fig.9(b), it is interested to find that the segmental result is greater than the ground truth in Fig.9(e). The small ships can be found  and at the meanwhile can be semantically segmented out from the sea area through the model of the DeepUNet. The result of U-Net (Fif.3(c)) is good as well. However, in the small areas especially near the boundary, there are a lot of misclassified pixels. This experiment shows that the U-Net cannot deal with the detailed areas and minor objects. In comparison, the famous SegNet does not get good performance in sea-land segmentation, though it ranks high in the ImageNet competition. There exist two reasons. First of all,  the sea-land images are usually of high-resolution and contain objects of various scales from small ships to large connected island. Secondly, the semantic content is different from that of natural images. The sea-land segmentation task is a pixel-level binary classification problem. It pays more attention to the connectivity of the area, which is traditionally solved by morphological methods. But it is a hard problem for CNN based methods. For example, the SegNet is base on the VGG16 net. Its architecture cannot afford deeper convolution layers for the complex connectivity problem. However, it can get high precision  segmental results along the boundary because of the encoder-decoder architecture. 

\begin{figure}[htbp]
\setlength{\abovecaptionskip}{0pt}
\setlength{\belowcaptionskip}{-10pt}
\centering
\includegraphics[width=.45\textwidth]{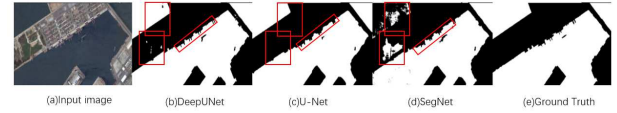}
\caption{the segmentation results of port predict by DeepUNet, U-Net, SegNet}
\label{fig9}
\end{figure}

For Fig.9 the evaluation table is listed in Tablb \uppercase\expandafter{\romannumeral4}. The indicators show that DeepUNet's OP is 3.65\% higher than U-Net and 3.26\% higher than SegNet. The F1-measure of DeepUNet is 4.8\% higher than U-Net and 1.92\% higher than SegNet.

\begin{table}[h]
\centering
\caption{the evaluation results of Fig.9}
\begin{tabular}{|c|c|c|c|c|c|}
\hline
Name & LP(\%) & LR(\%) & OP(\%) & OR(\%) & F1(\%) \\
\hline
DeepUNet & \textbf{96.02} & \textbf{96.02} & \textbf{98.14} & \textbf{98.14} & \textbf{96.02} \\
\hline
U-Net & 94.91 & 99.81 & 97.71 & 97.71 & 97.30 \\
\hline
SegNet & 80.99 & 95.33 & 88.85 & 88.85 & 87.57 \\
\hline
\end{tabular}
\label{table4}
\end{table}

Fig.10 shows a special case. It is an remote sensing image that contains facilities  on the sea. The ocean area is clear without waves and other interference factors. We use this image to further test the DeepUNet's performance when facing various boundary and small objects.  The result of the DeepUNet are almost correct. In comparison, the U-Net cannot deal with the detailed areas and the SegNet fails to distinct all the land areas. It is obvious that  ships in the sea can be  segmented out through the DeepUNet. The detailed of indicators are shown in TABLE \uppercase\expandafter{\romannumeral5}.

\begin{figure}[h]
\setlength{\abovecaptionskip}{0pt}
\setlength{\belowcaptionskip}{-10pt}
\centering
\includegraphics[width=.45\textwidth]{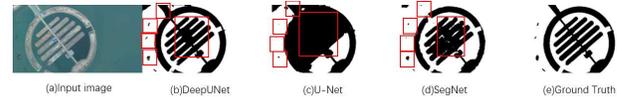}
\caption{the segmentiation result of  the building which is Complex structure}
\label{fig10}
\end{figure}

\begin{table}[h]
\centering
\caption{the evaluation results of Fig.10}
\begin{tabular}{|c|c|c|c|c|c|}
\hline
Name & LP(\%) & LR(\%) & OP(\%) & OR(\%) & F1(\%) \\
\hline
DeepUNet & \textbf{91.73} & \textbf{99.35} & \textbf{97.50} & \textbf{97.50} & \textbf{95.39} \\
\hline
U-Net & 64.74 & 98.94 & 85.70 & 85.70 & 78.27 \\
\hline
SegNet & 89.30 & 91.44 & 94.92 & 94.92 & 90.35 \\
\hline
\end{tabular}
\label{table5}
\end{table}

More results can be found in Fig.1. On all the 85 testing images, the overall LP, LR, OP, OR and F1 are listed in TABLE \uppercase\expandafter{\romannumeral6}. Both the indicators of the DeepUNet are higher than the that of the other two networks. The SeNet also promoted a sea-land segmentation architecture. It chooses a multi-task way to combine the image segmentation and edge detection to get better results than original DeconvNet. The SegNet is very like the DeconvNet in structure except  that it optimizes the encoding and decoding strategy. Moreover, the work [15] demonstrated that the SegNet has better performance than DeconvNet in image segmentation on the cityscape dataset. In our experiment, we did not compare the DeepUNet to the SeNet directly, since we neither had their datasets nor implemented their architecture. Instead, we compare our network with the SegNet which is better than DeconvNet. It is indicated that the DeepUNet outperforms the SegNet or DeconvNet a lot.  The contribution of the DeepUNet is to provide a more concise and elegant network structure. It is in fact not conflict with the multi-task method that the SeNet introduced.

\begin{table}[h]
\centering
\caption{the evaluation results of all the 85 testing images}
\begin{tabular}{|c|c|c|c|c|c|}
\hline
Name & LP(\%) & LR(\%) & OP(\%) & OR(\%) & F1(\%) \\
\hline
DeepUNet & \textbf{98.58} & \textbf{98.91} & \textbf{99.04} & \textbf{99.04} & \textbf{98.74} \\
\hline
U-Net & 96.68 & 97.42 & 97.57 & 97.57 & 97.05 \\
\hline
SegNet & 97.52 & 96.50 & 97.81 & 97.81 & 97.01 \\
\hline
\end{tabular}
\label{table6}
\end{table}

\section{Conclusion and Future works}
In this paper, we designed a very elegant symmetric neural network named DeepUNet for pixel-level sea-land segmentation. DeepUNet is an end-to-end fully convolutional network with two other kinds of short connections. We call them U connections and Plus connections.  We specifically designed the DownBlock structure and the UpBlock structure to adopt these connections. \\

To verify the network architecture, we collected a set of remote sensing Sea-land data RGB image sets from Google-Earth. And, we manually labeled the ground truth. On the collected dataset, we compare the DeepUnet with the SeNet and the SegNet. Experiments results show that the proposed DeepUNet outperformed the other networks significantly. \\

In the future, we intend to combine the multi-task learning technique to our architecture to further enhance accuracy.\\


\end{document}